\documentclass[10pt,twocolumn,letterpaper]{article}

\usepackage{cvpr}
\usepackage{times}
\usepackage{epsfig}
\usepackage{graphicx}
\usepackage{amsmath}
\usepackage{amssymb}
\usepackage{enumitem}
\usepackage{booktabs}
\usepackage{algorithm}
\usepackage{algpseudocode}
\usepackage{xcolor}

\usepackage[pagebackref=true,breaklinks=true,letterpaper=true,colorlinks,bookmarks=false]{hyperref}

\cvprfinalcopy %

\def\cvprPaperID{6165} %

\ifcvprfinal\fi
\begin{document}
	
	\title{Optical Non-Line-of-Sight Physics-based 3D Human Pose Estimation}
	
	\author{Mariko Isogawa, Ye Yuan, Matthew O'Toole, Kris Kitani\\
		Carnegie Mellon University\\
	}

	\maketitle
	
	\begin{abstract}
		We describe a method for 3D human pose estimation from transient images (\ie, a 3D spatio-temporal histogram of photons) acquired by an optical non-line-of-sight (NLOS) imaging system. Our method can perceive 3D human pose by `looking around corners' through the use of light indirectly reflected by the environment. We bring together a diverse set of technologies from NLOS imaging, human pose estimation and deep reinforcement learning to construct an end-to-end data processing pipeline that converts a raw stream of photon measurements into a full 3D human pose sequence estimate. Our contributions are the design of data representation process which includes (1) a learnable inverse point spread function (PSF) to convert raw transient images into a deep feature vector; (2) a neural humanoid control policy conditioned on the transient image feature and learned from interactions with a physics simulator; and (3) a data synthesis and augmentation strategy based on depth data that can be transferred to a real-world NLOS imaging system. Our preliminary experiments suggest that our method is able to generalize to real-world NLOS measurement to estimate physically-valid 3D human poses.\footnote{Project page: \href{https://marikoisogawa.github.io/project/nlos_pose}{https://marikoisogawa.github.io/project/nlos\_pose}}
	\end{abstract}

	\vspace{-5mm}
	\section{Introduction}
	\label{sec:introduction}
	
	Our goal is to develop a vision-based human pose estimation technique capable of inferring the 3D pose sequence of a person without direct line of sight from the sensing device (see Fig.~\ref{fig:teaser}). To this end, this work brings together, for the first time, state-of-the-art techniques in computational imaging, human pose estimation, humanoid control, and deep reinforcement learning -- all under one system architecture. From computational imaging, we use volumetric reconstruction algorithms that efficiently process transient measurements (a 3D spatio-temporal histogram of photons) captured from a non-line-of-sight (NLOS) imaging system. From vision-based pose estimation and humanoid control, we make use of deep reinforcement learning in a physics simulator to learn an image feature-conditioned 3D human pose policy that abides by the laws of physics. All components are integrated into a single end-to-end learnable architecture which takes the raw transient images, converts them to a deep feature representation using a learnable inverted point spread function (PSF), and uses a neural humanoid control policy conditioned on the deep feature to estimate physically valid 3D human poses. 
	In short, our proposed end-to-end system models the complete flow of information from the photon-level all the way to the high-level cognitive task of 3D human pose estimation.
	
	In order to capture the actions of a person around a corner, we employ a NLOS imaging technique using a pulsed laser and a time-of-flight sensor \cite{Velten2012}. The imaging process involves sending light pulses towards a visible surface, and measuring the travel time of light reflected back in response; this represents a \emph{``transient image''} \cite{Velten2013,Gkioulekas2015,O'Toole2017}. The transient image used here is a 3D volume that encodes the arrival of individual photons over a spatial region (first two dimensions) and over time (third dimension); see Fig.~\ref{fig:teaser} for a sequence of transient images. Optical NLOS imaging at visible or near-IR wavelengths offers several advantages, such as the ability to image at long-ranges~\cite{Chan2017}, reconstruct surfaces at micron-scales~\cite{Xin2019}, and robustly perform imaging for multiple objects with different BRDFs~\cite{Lindell2019}.
	
	While NLOS imaging gives us the ability to look around corners, the raw transient measurements acquired by a NLOS system have several unique properties that make 3D human pose estimation difficult. In order to attain fast image acquisition necessary for 3D human pose understanding, existing NLOS solutions limit the resolution of the transient image in both space and time~\cite{Lindell2019}. This makes it difficult to capture small shape details and fast motion, both of which are important for estimating a human pose sequence. Furthermore, due to the light lost after multiple scattering events, very few photons reach the sensor and the acquired transient image can therefore be very noisy. All of these characteristics make it very challenging to estimate 3D human pose directly from the transient image.
	
	Due to the noisy nature of transient images, it is important to make use of prior information about the body and its dynamics to help solve the 3D human pose estimation problem. In particular, we have prior knowledge about the structure of the human body (\eg, arms, head, and legs are connected to the torso) and we also know that human pose must abide by the laws of physics. Based on this prior knowledge, we make use of a parametrized humanoid model and a physics-based deep reinforcement learning approach to estimate 3D human pose from a sequence of transient images. Inspired by \cite{Yuan2019}, we first define a Markov Decision process (MDP) where the state is defined as the current pose of a humanoid model and a sequence of transient images, the action is defined as the target joint angles of a humanoid, the environment is a physics simulator (\ie, MuJoCo~\cite{mujoco}) and the reward function is an imitation-based objective function \cite{Peng2018_1}. By solving the MDP via reinforcement learning, we are able to learn an optimal policy that maps states to actions. In our scenario, the policy function encodes how the humanoid should move such that it best explains the 3D pose information captured in the transient image sequence.
	
	The deep neural network design used to implement the policy of the MDP is a critical representational element of the system, as it must have the computational components necessary to map a 3D transient image to a vector of humanoid joint angles. 
	Inspired by recent NLOS imaging techniques, we model our network based on existing reconstruction algorithms used to process transient images, and introduce P2PSF Net: a neural network that learns ``corrections'' that improve the NLOS image reconstruction process. A deep feature computed by the P2PSF Net is then passed to a bi-directional recurrent neural architecture to ensure that the output of the policy (the next joint position) follows the laws of motion and matches demonstrated human pose dynamics. The ambitious goals of the policy is to estimate fine-scale 3D human pose from transient images and overcome the resolution limits and noisy characteristics of NLOS imaging.
	
	As mentioned above, the MDP formulation for human pose estimation makes use of an imitation-based reward function which requires a large amount of annotated data in the form of transient images and a corresponding 3D human pose sequence. As a practical matter, current NLOS imaging systems are quite costly to build and non-trivial to calibrate. As such, the need for a large amount of annotated data and the cost of collecting such data are at odds with each other. While we do believe that these practical hurdles will be resolved through new innovations over time, for this work, we address them by synthesizing pseudo-transient images using the 3D body volume recovered with a depth camera. To learn our MDP policy, we rely solely on this synthetic data synchronized with ground truth 3D human pose acquired by a motion capture system. 
	
	Synthesizing transient images from depth maps for training can create a large domain gap between the training and test data distributions. 
	To ensure that we minimize the size of this domain gap,
	it is critical to implement principled data augmentation techniques based on the physics of the imaging process, \eg, reproducing the noise characteristics associated with real transient images. We introduce several augmentation techniques 
	to make the 3D human pose estimation policy able to handle real-world NLOS data.
	
	To summarize, our contributions are as follows: 
	\\\noindent(1) We are the first to propose an end-to-end data processing pipeline that models information flow from optical NLOS transient measurements to 3D human pose estimation.
	\\\noindent(2) We present a novel policy network architecture that integrates a inverse point spread function (PSF), transient images, and human pose as one deep neural network.
	\\\noindent(3) We describe a method for generating pseudo (\ie, synthetic) transient images from depth images. We propose several critical processes for augmenting pseudo-transient images to help the learned 3D human pose policy to generalize better to real-world data. 
	\\\noindent(4) We provide extensive experimentation with both synthetic and real transient images and show that our model outperforms other baseline methods.

	\section{Related Work}
	\label{sec:related_work}
	
	\noindent
	\textbf{NLOS Imaging and Its Applications.}
	NLOS imaging has received significant attention recently \cite{Adib2015,Bouman2017,Chan2017,Heide2014,Klein2016,Klein2017,Li2019,lindell2019acoustic,Lindell2019, o2018confocal, Velten2012,Maeda2019,redo2016terahertz,Tancik2018,Tsai2019,Xin2019,Zhao2018}.  This includes a wide variety of solutions, in terms of hardware systems and reconstruction algorithms, that operate in different parts of the electromagnetic spectrum. Most NLOS solutions work with visible (380-740~nm) or near-IR (740-1500~nm) light~\cite{Chan2017,Lindell2019,Xin2019}; we refer to these as \emph{optical} NLOS systems.  Others choose to operate at longer wavelengths, including long-wave IR (8-15~um)~\cite{Maeda2019}, terahertz (millimeter scale)~\cite{redo2016terahertz}, WiFi and radio frequency (RF) (centimeter scale)~\cite{Li2019}, or even sound waves~\cite{lindell2019acoustic}.
	
\begin{figure}
		\begin{center}
			\includegraphics[width=1\linewidth]{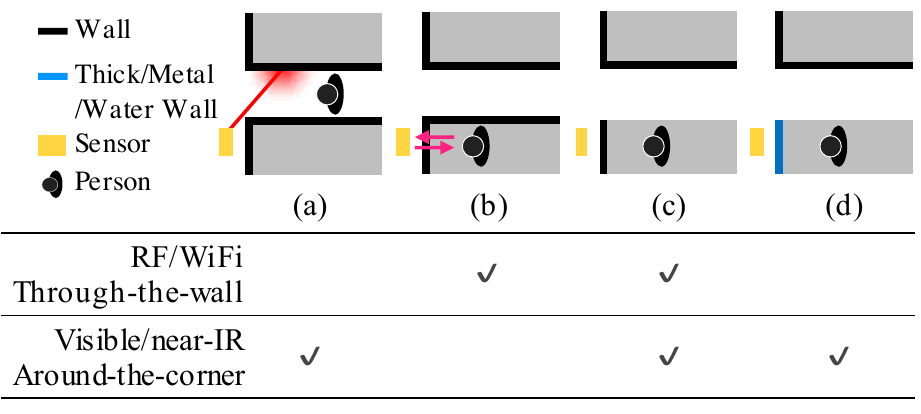}
		\end{center}
		\vspace{-2mm}
		\caption{Overview of WiFi/RF (\ie, through-the-wall) and visible/near-IR (\ie, around-the-corner) NLOS human pose estimation. \textbf{(a)} The person is hidden from view in a nearby corridor, but seen indirectly by reflecting light off of a visible surface.  \textbf{(b)} The person is hidden in an enclosed room, which can only be accessed with WiFi/RF that pass through walls.  \textbf{(c)} The person is within a partially-enclosed space that can be accessed through either method.  \textbf{(d)} The walls are either too thick or made from a material that prevents WiFi and RF signals from passing through.  
		}
		\label{fig:diffrences_rf}
	\end{figure}

	Operating in a specific part of the spectrum affects the nature of the NLOS signal that can be used for pose estimation.
	For example, RF or WiFi signals enable through-the-wall pose estimation~\cite{Adib2015,Li2019,Zhao2018}, since longer electromagnetic waves tend to pass through objects; however, as highlighted in Fig.~\ref{fig:diffrences_rf}, this also comes at a number of fundamental limitations when estimating pose dynamics: the spatial and angular resolution must be limited, making it hard to distinguish small objects or motions; objects tend to behave specularly at long wavelengths, presenting a challenge when attempting to form images; RF signals can be highly attenuated when passing through thick objects; and RF signals are blocked entirely when passing through other surfaces, like metal or water. %
	As a result, they cannot always be applied to scenes shown in Fig.~\ref{fig:diffrences_rf} (a), (c) and (d).
	
	Rather than using RF signals to attempt imaging through walls, most NLOS works choose to operate in the optical domain and image \emph{around} corners.  This is done by reflecting light off of walls to indirectly illuminate a hidden scene, and measuring the time light takes to return to the wall with transient sensors, \eg, single photon avalanche diodes (SPADs). Amongst these works, many attempt to reconstruct 3D volumes~\cite{Heide2014,Lindell2019, o2018confocal, Velten2012} where each voxel represents the scene's reflectivity at a 3D location; or compute a surface representation~\cite{Tsai2019,Xin2019}, offering the potential for more accurate reconstructions.  Even regular cameras can be used to track a person walking around in a room under either passive~\cite{Bouman2017} or active~\cite{Klein2016,Klein2017,Tancik2018} illumination.
	
	Data-driven methods for optical NLOS imaging have also received some attention, \eg, to discretely classify the pose of a human from raw transient measurements~\cite{Satat:17} or form NLOS images with a regular camera~\cite{Tancik2018}.  Network architectures have also been proposed for denoising transient measurements from SPADs, in the context of line-of-sight imaging~\cite{Lindell2018}.
	However, there has been no prior work addressing the task of learning full physics-based 3D human pose estimation in a NLOS context. 
	
	\noindent
	\textbf{Physics Based 3D Human Pose Estimation.}
	Enforcing physics-based on constraints on human pose dynamics is used commonly for simulated humanoid control \cite{Peng2018_1, Peng2017, Peng2018_2,Yuan2018, Yuan2019}. These methods use deep reinforcement learning (DeepRL) to learn control policies that can reproduce humanoid motion inside a physics simulator. Among these methods, \cite{Yuan2018, Yuan2019} use the optical flow of egocentric videos as additional inputs to the motion policy to estimate physically-valid human poses. We use a similar RL-based pose estimation framework proposed in~\cite{Yuan2019} for its ability to estimate physically-accurate human poses with only limited amounts of visually information. However, the transient images that we use as input present many new challenges when learning the policy, which we address in this paper.

	\begin{figure*}
		\begin{center}
			\includegraphics[width=1.0\textwidth]{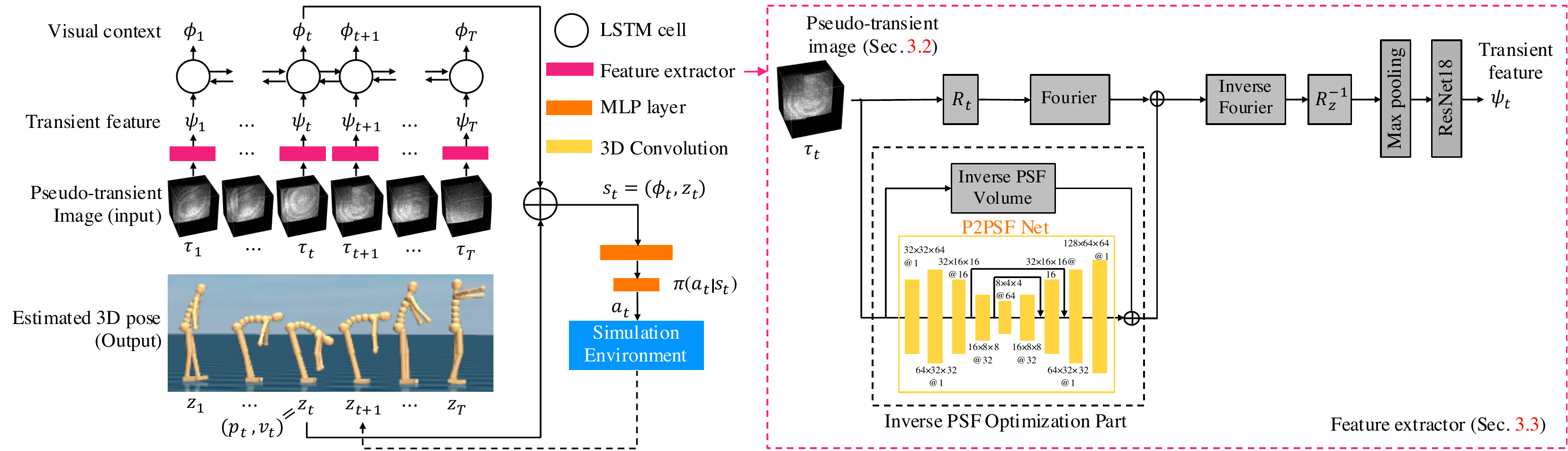}
		\end{center}
		\caption{Our DeepRL based photons to 3D human pose estimation framework under the laws of physics.}
		\label{fig:models}
	\end{figure*}

	\section{Methodology}
	
	There are three key ideas behind our method. First, to model physically-valid human pose, we use a humanoid model inside a physics simulator and model hidden 3D human pose as the result of a humanoid control policy conditioned on the input transient images (Sec.~\ref{sec:framework}). Second, to make our model applicable to real captured data, we synthesize ``pseudo-transient'' images as training data which replicates the visual characteristics of real transient measurements (Sec.~\ref{sec:pseudo_transient}). Third, to enhance our pose estimation, we further introduce P2PSF Net that improves the NLOS imaging process (Sec.~\ref{sec:transient_to_pose}).

	\subsection{Physics-based 3D Pose Estimation Framework}
	\label{sec:framework}
	Given a sequence of transient images $\tau_{1:T}$, the goal of our physics-based pose estimation pipeline is to predict a physically-valid pose sequence $p_{1:T}$. As shown in Fig.~\ref{fig:models}, the pipeline contains two main parts: (1) We use a feature extractor consisting of the NLOS imaging pipeline and P2PSF Net to extract transient features $\psi_{1:T}$ from the transient images $\tau_{1:T}$; (2) We use a humanoid policy conditioned on the transient features $\psi_{1:T}$ to control the humanoid inside a physics simulator and generate the pose sequence $p_{1:T}$ underlying the transient images. In this section, we will focus on part 2, \emph{i.e.,} humanoid control, and leave details on how to extract transient features (part 1) to Sec.~\ref{sec:pseudo_transient} and \ref{sec:transient_to_pose}.
	
	Following~\cite{Yuan2019}, we formalize the task of estimating a pose sequence $p_{1:T}$ from a transient image sequence $\tau_{1:T}$ with a Markov Decision process~(MDP). The MDP is defined by a tuple $\mathcal{M} = (S, A, P, R, \gamma)$ of states, actions, transition dynamics, a reward function, and a discount factor. 
	As Fig.~\ref{fig:models} shows, at each time step, the humanoid agent samples an action $a_t$ from a policy $\pi (a_t | s_t)$ whose input state $s_t$ contains both the visual context $\phi_t$ (defined later)
	and the humanoid state $z_t$ (\emph{i.e.}, position and velocity of the human model joints). Next, the environment generates the next state $s_{t+1}$ through physics simulation and gives the agent a reward $r_t$ based on how well the humanoid's 3D pose aligns with the ground-truth.
	To solve this MDP, we apply the PPO policy gradient method~\cite{Schulman2017} to obtain the optimal policy $\pi^*$ that maximizes the expected discounted return $\mathbb{E} [ \sum^T_{t=1} \gamma^{t-1} r_t ]$. At test time, starting from some initial state $s_1$, we roll out the policy $\pi^*$ to generate state sequence $s_{1:T}$, from which we extract the output pose sequence $p_{1:T}$. In the following, we will discuss the details of each component.
	
	\noindent
	\textbf{State} 
	$s_t$ consists of the state of the humanoid $z_t$ and the visual context $\phi_t$. $z_t$ consists of the pose $p_t$ (root position/orientation and joint angles) and velocity $v_t$ (root linear/angular velocities and joint velocities). The visual context $\phi_t$ is extracted from the transient feature $\psi_{1:T}$ with a bi-directional LSTM (BiLSTM) as shown in Fig.~\ref{fig:models}. During training, we set the starting state $z_1$ to the ground-truth $\hat{z}_1$. Since we have no access to the ground truth at test time, we learn a regressor $\mathcal{F}$ that maps the visual context $\phi_t$ to its corresponding state $z_t$.

	\noindent
	\textbf{Action} 
	$a_t$ specifies the target joint angles of the Proportional-Derivative~(PD) controller for each degree of freedom~(DoF) of the humanoid joints except for the root. The joint torques are computed based on the stable PD controllers in~\cite{Tan2011} using the specified target joint angles $a_t$.

	\noindent
	\textbf{Policy} 
	$\pi_{\theta}(a_t | s_t)$ is represented by a Gaussian distribution $\mathcal{N}(\mu; \Sigma)$ with a fixed diagonal covariance matrix treated as hyperparameters. We use a $\theta$-parametrized multilayer perceptron~(MLP) with two hidden layers (300, 200) and ReLU activations to model the mapping from $s_t$ to the mean $\mu_t$.

	\noindent
	\textbf{Reward function.} 
	To encourage the policy to output a pose sequence $p_{1:T}$ that matches the ground-truth $\hat{p}_{1:T}$, we use the reward function proposed in~\cite{Yuan2019}. The specific design of the reward function is given in Appendix~\ref{sec:reward}.

	\subsection{Pseudo-Transient Image}
	\label{sec:pseudo_transient}
	
	Before introducing our transient data synthesis from depth maps (Sec.~\ref{sec:our_data_synthesis}), we provide background on the NLOS imaging procedure used in this work in Sec.~\ref{sec:conventional_lct}. 
	
	\subsubsection{Background: Confocal NLOS Imaging}
	\label{sec:conventional_lct}
	
	Confocal NLOS imaging refers to a specialized raster-scanning procedure for capturing transient measurements~\cite{o2018confocal}.  Measurements are the result of illuminating and imaging a common point $(x',y')$ on a visible surface.  After illuminating this point with a pulse laser, light scatters to hidden regions of an environment and returns back to the wall at a later instant of time.  A SPAD measures the transient response at this same point, represented as a histogram of photon arrival times~\cite{O'Toole2017}.  This procedure is repeated for a uniform and planar 2D grid of points across the surface; the collection of transients is stored as a 3D transient image $\tau(x,y,t)$.
	
	The objective of NLOS imaging is to convert this 3D transient $\tau(x',y',t)$ into a discretized reconstruction volume $\rho(x, y, z)$, which represents the reflectivity at every point $(x,y,z)$ in space.  The presence of an object in voxel $(x,y,z)$ produces a non-zero reflectance value $\rho(x,y,z)$, where the wall is located at $z=0$.%
	
	After resampling the transient measurements $\tau$ along the time dimension ($\Tilde{\tau}=R_t\{\tau\}$) and resampling the reconstruction volume $\rho$ along the depth dimension ($\Tilde{\rho}=R_z\{\rho\}$), the forward image formation model for confocal NLOS images becomes a simple 3D convolution operation~\cite{o2018confocal}:
	\begin{equation}
	\Tilde{\tau} = h * \Tilde{\rho},
	\label{eq:conv}
	\end{equation}
	where the hidden volume $\Tilde{\rho}$ is convolved with a known 3D point spread function (PSF) represented by $h$.  This PSF $h$ describes the transient response of a single scatterer.
	
	Equivalently, we can rewrite this convolution in matrix-vector form as follows:
	\begin{equation}
	\tau = 
	R_t^{-1} F^{-1} \hat{H} F R_z \rho,
	\label{eq:lct_discrete}
	\end{equation}
	where we vectorize volumes $\tau$ and $\rho$.  The matrix $F$ represents a 3D discrete Fourier transform and $\hat{H}$ is a diagonal matrix representing the Fourier transform of the PSF $h$.
	
	The NLOS imaging process reconstructs a 3D volume $\rho_\ast$ from a transient image $\tau$ by
	inverting Eq.~\ref{eq:lct_discrete} and solving a 3D deconvolution procedure (\eg, using the Wiener filter):
	\begin{equation}
	\rho_{*} = 
	R_z^{-1} F^{-1} \underbrace{\left[\frac{\hat{H}^*}{|\hat{H}|^2 + \frac{1}{\alpha}}\right]}_{\textbf{the inverse PSF}} F R_t \tau\,,
	\label{eq:lct_final}
	\end{equation}
	where a user-defined parameter $\alpha$ controls how sensitive the inverse PSF is to noise.

	\subsubsection{Pseudo-Transient Image Synthesis}
	\label{sec:our_data_synthesis}
	
	We propose a procedure for synthesizing pseudo-transient images to generate annotated training data for our pose estimator.  This involves using a motion capture (MoCap) system to obtain 3D ground truth human pose, which is synchronized with a depth camera to capture depth images used for synthesizing pseudo-transient images.
	
	Given a depth map $d(x, y)$, we compute a corresponding synthesized reflectance volume for our scene: $\rho_{s}(x, y, z) = a$ when $z=d(x, y)$, and zero otherwise.  The scalar $a=100$ is a constant value representing the amount of light reflected by voxels in the volume.\footnote{Though we assume uniform reflectance in this paper, an intensity image could be used to encode non-uniform reflectance information.}  We then convert the synthesized volume $\rho_{s}$ into transient measurements $\tau$ by using the image formation model described in Eq.~\ref{eq:lct_discrete}.

	Note, however, that the image formation model is currently incomplete, since it does not model all characteristics of real-world transient measurements.  For dynamic scenes (\eg, a person walking around in a room), acquisition times must be short.  Raster scanning transient measurements, therefore, is affected by motion, and the number of sampled points on the wall may be limited.  The sensors used for NLOS imaging (\eg, SPADs) are also affected by different types of sensor noise.%
	
	To close the domain gap that exists between the synthesized data and the real transient measurements, we introduce Poisson noise, temporal blur, temporal shifts, and temporal re-sampling into the image formation model.
	
	\noindent
	\textbf{Poisson noise.} 
	At low light levels, the number of photons detected by a SPAD approximately follow Poisson noise characteristics~\cite{O'Toole2017}.  We therefore apply Poisson noise to our synthetic transient measurements, as a post-process operation on Eq.~\ref{eq:lct_discrete}.

	\noindent
	\textbf{Temporal blur.} 
	The temporal profile of a transient measurement is affected by jitter and the shape of the laser pulse.  The result is that the transient measurements become blurred in the time domain. Before the Poisson noise step, we introduce temporal blur by convolving the transients measurements with a Gaussian~\cite{O'Toole2017}.  The standard deviation of the Gaussian is characterized by its full width at half maximum (FWHM), which is reported to be 70 picoseconds for the real-world NLOS data used in this work~\cite{Lindell2019}.

	\noindent
	\textbf{Temporal shift. } 
	The arrival times of photons are correlated with the distance of the hidden person from the wall.  We augment our data by biasing the depth values by a constant amount $d$ prior to synthesizing pseudo-transient images, which temporally shifts the transient images.  In our experiments, we make use of five levels of shifts, augmenting the training data up to a factor of five.

	\noindent
	\textbf{Temporal re-sampling. } 
	Confocal NLOS imaging requires raster scanning a visible surface point by point.  The mechanics of current systems impose limits on the speed of this raster scanning procedure.  For example, we make use of NLOS data sampled at $32 \times 32$ locations at a frame rate of only $4$ Hz~\cite{Lindell2019}.  Since points and corresponding transients are scanned sequentially, performance of pose estimation can be affected by fast moving body parts.

	We propose a procedure to temporally re-sample transient data.  First, to simulate the raster scanning procedure, we capture depth maps at $30$ Hz, convert them to pseudo-transient measurements, and simulate raster-scanned measurements by down-sampling the result to $4$ Hz.  Second, we up-sample the pseudo-transient data from $4$ Hz back to $30$ Hz.  Each transient image is a collection of transients scanned from time $t_k$ to $t_k + \frac{1}{4}$, where $t_k = \frac{k}{4}$ represents the start time of the $k^\text{th}$ frame (in seconds).  To generate a $30$ Hz transient sequence, we simply assemble transients captured within the same time range, but set the start time of the $k^\text{th}$ frame to $t_k = \frac{k}{30}$. For real transient images used at test time, we also perform above procedure to up-sample them to 30 Hz to be compatible with the humanoid policy.

	\subsection{P2PSF Net: Photon to Inverse PSF Network}
	\label{sec:transient_to_pose}
	
	To predict 3D human pose from transient images, as discussed in Sec.~\ref{sec:framework}, Our DeepRL based framework needs a feature extractor to obtain transient feature $\psi_t$ from transient image $\tau_t$. We model the feature extractor by incorporating aspects of the confocal NLOS imaging process (described in Sec.~\ref{sec:conventional_lct}) into our feature extractor network.
	Note that learning to estimate 3D human pose from transient images is challenging for two key reasons: (1) the transient images are noisy, have low spatial resolution, and are recorded at slow frame rates; (2) the inverse PSF, labeled in Eq.~\ref{eq:lct_final}, makes several simplifying assumptions about the NLOS imaging process.

	To overcome these challenges, we propose P2PSF Net: the photon-to-inverse-PSF network.  Its objective is to adjust the inverse PSF volume used in the confocal NLOS reconstruction process (\eg, to help handle realistic sensor noise, calibration errors, motion during the acquisition period). 
	P2PSF Net is a 3D volume-to-volume network with nine 3D convolution layers (see Fig.~\ref{fig:models}). It also has two residual connections, which are inspired by past works~\cite{resnet,unet}. Given a transient image of resolution $(x,y,t)=32\times32\times64$, the network outputs a volume $(x,y,d)=128\times64\times64$, matching the size of the inverse PSF volume used in confocal NLOS imaging.

	To extract transient features $\psi_t$ , we first reconstruct the reflectance volume $\rho_\ast$ following the procedure used in confocal NLOS imaging (Eq.~\ref{eq:lct_final}) with the modification that we add the output of P2PSF Net as a corrective volume to the inverted PSF volume.
	Given the reconstructed reflectance volume $\rho_\ast$, we then generate a 2D heat map by applying a single max pooling layer across the depth axis of the volume.
	The 2D heat map is then passed to a ResNet-18~\cite{resnet} pretrained on ImageNet~\cite{imagenet} to extract $\psi_t$.

	\begin{figure*}
		\begin{center}
			\includegraphics[width=1.0\textwidth]{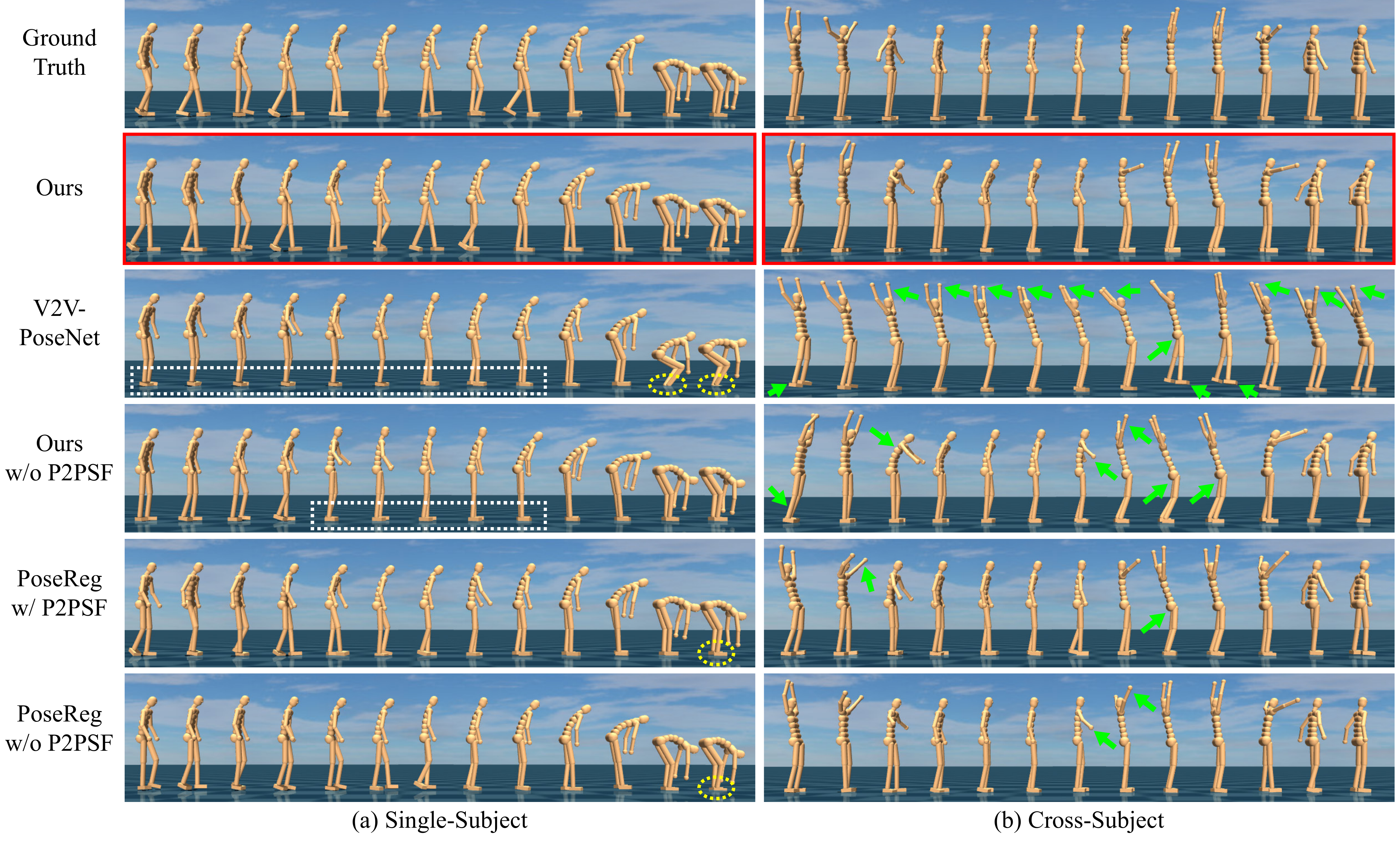}
		\end{center}
		\vspace{-15pt}
		\caption{Qualitative results for \textbf{(a)} single-subject and \textbf{(b)} cross-subject pose estimation.}
		\label{fig:single_and_cross}
	\end{figure*}

	\section{Experimental Settings}
	\label{sec:experimental_settings}

	\subsection{Datasets}
	\label{subsec:datasets}
	
	\noindent
	\textbf{Pseudo-Transient Image Dataset.} 
	We generated a large set of synthetic pseudo-transient images from MoCap data. The dataset is $1$ hour long and has 103200 frames with 30 FPS. It consists of five subjects performing various complex motions: walking, jumping, turning, bending backward/forward, rotating, and transitioning between all of these motions. Our method does not require segmenting the pose sequences or labeling the actions. 
	
	\noindent
	\textbf{Real Captured Transient Image Dataset.} 
	To further showcase our method’s applicability, we also test our method on actual transient images~\cite{Lindell2019}. This is a challenging dataset since it contains a different subject and the data acquisition conditions are different compared to the training data.  This test data also contains significant noise, samples a sparse grid ($32 \times 32$), and records measurements at a low frame-rate (\ie, $4$ Hz). The person is wearing a retroreflective tracksuit to increase the light signal used in NLOS imaging. Through our temporal re-sampling procedure (Sec.~\ref{sec:our_data_synthesis}), we up-sampled this data to 30 FPS and 1000 frames. Since this dataset has no pose ground truth, we extract 2D keypoint ground truth with AlphaPose~\cite{Xiu2018} using a third-person RGB camera.

	\subsection{Baseline Methods}
	\label{subsec:baselines}
	\begin{enumerate}[leftmargin=*]
		\setlength\itemsep{-3pt}
		\item
		\textbf{V2V-PoseNet~\cite{Moon2018}:}
		Since there is no existing work on optical NLOS human pose estimation from transient images, we compare against the state-of-the-art method, V2V-PoseNet, for depth volume-based pose estimation~\cite{Moon2018}. We train V2V-PoseNet using the reconstructed depth volume $\rho_\ast$ computed from our synthetic pseudo-transient images using the regular NLOS imaging process (without P2PSF Net) defined in Eq.~\ref{eq:lct_final}.
		
		\item
		\textbf{PoseReg:}
		To investigate the effect of physics inside our physics-based pose estimation framework, we compare our method a regression-based method, which directly maps the visual context $\phi_t$ to the humanoid state $z_t$ without using any physics. We integrate root linear/angular velocities to generate global positions and orientations of the pose sequence. 
		
		\item
		\textbf{PoseReg w/o P2PSF:} 
		a variant of PoseReg that does not use proposed P2PSF Net in the NLOS imaging process.
		
		\item
		\textbf{Ours w/o P2PSF:} 
		a variant of our method that does not use proposed P2PSF Net.
		
	\end{enumerate}

	\subsection{Evaluation Metrics}
	\label{subsec:metrics}
	
	We use the following metrics to evaluate both the accuracy and physical correctness of each method:
	
	\begin{enumerate}[leftmargin=*]
		\setlength\itemsep{-3pt}
		\item
		\textbf{Mean Per-Joint Position Error (MPJPE)}:
		a pose-based metric that measures the Euclidean distance (measured in millimeters) between the ground truth and prediction for a joint defined as $\frac{1}{TJ}\Sigma^{T}_{t=1}\Sigma^{J}_{j=1} ||(x^j_t - x^{root}_t) - (\hat{x}^j_t - \hat{x}^{root}_t)||_2$, where $x^j_t$ is the $j^\text{th}$ joint position of estimated pose and $\hat{x}^j_t$ is the ground truth. $x^{root}_t$ and $\hat{x}^{root}_t$ represent root joint positions of the estimate and the ground truth.
		
		\item
		\textbf{2D Keypoint Error (}$\mathbf{E}_{\textrm{key}}$\textbf{):} 
		A pose-based metric used for real captured dataset, calculated as $\frac{1}{TJ}\Sigma^{T}_{t=1}\Sigma^{J}_{j=1} ||y^j_t - \hat{y}^j_t||_2$. Here, $y^j_t$ is the $j^\text{th}$ 2D keypoint of estimated pose and $\hat{y}^j_t$ is the ground truth. For both estimated and ground-truth, we set the hip keypoint as the origin and scale the coordinates to make the height between shoulder and hip equal to $0.5$.
		
		\item
		\textbf{Velocity Error (}$\mathbf{E}_{\textrm{vel}}$\textbf{):} 
		A physics-based metric that measures the Euclidean distance between the generated velocity sequence $v_{1:T}$ and the ground-truth $\hat{v}_{1:T}$, calculated as $\frac{1}{T}\Sigma^{T}_{t=1} ||v_t - \hat{v}_t||_2$.
		
		\item
		\textbf{Average Acceleration (}$\mathbf{A}_{\textrm{accl}}$\textbf{)}: 
		A physics-based metric that uses the average magnitude of joint accelerations to measure the smoothness of the pose sequence, calculated as $\frac{1}{TN}\Sigma^{T}_{t=1} ||\dot{v}^{t}||_1$ where $\dot{v}_t$ denotes joint accelerations and $N$ is the number of actuated DoFs.
	\end{enumerate}

	\subsection{Implementation Details}
	\label{subsec:implementation_details}
	
	\noindent
	\textbf{Simulation Environment. } 
	We use MuJoCo~\cite{mujoco} as the physics simulator for a humanoid that consists of 58 DoFs and 21 rigid bodies. 
	We use stable PD controllers~\cite{Tan2011} to compute joint torques. The gains $k_p$ ranges from 50 to 500, where joints such as legs and spine have larger gains while arms and head have smaller gains; $k_d$ is set to 0.1$k_p$. 
	
	\noindent
	\textbf{Networks and Training.} 
	We set the reward weights $(w_q, w_e, w_p, w_v)$ to (0.5, 0.3, 0.1, 0.1). We use PPO~\cite{Schulman2017} with a clipping epsilon of 0.2 for policy optimization. The discount factor $\gamma$ is 0.95. We use Adam~\cite{adam} to optimize the policy and value function with learning rate 5e-5. The policy typically converges after 3k iterations, which takes about a day on a GeForce RTX 2080 Ti.

	\begin{table*}
		\footnotesize
		\mbox{}\hfill
		\begin{minipage}[t]{.7\linewidth}
			\centering
			\caption{Results on single (left), cross-subject (middle), and real captured data (right).}
			\begin{tabular}[t]{@{\hskip 1mm}lcccc@{\hskip -1mm}ccc@{\hskip 1mm}ccc@{\hskip 1mm}}
				\toprule
				& \multicolumn{3}{c}{Single Subject} & & \multicolumn{3}{c}{Cross Subject} & & \multicolumn{1}{c}{Real Data}\\ \cmidrule{2-4} \cmidrule{6-8} \cmidrule{10-10}
				Method & MPJPE $\downarrow$ & $\mathbf{E}_{\textrm{vel}}\downarrow$ & $\mathbf{A}_{\textrm{accl}}\downarrow$ & &  MPJPE $\downarrow$ & $\mathbf{E}_{\textrm{vel}}\downarrow$ & $\mathbf{A}_{\textrm{accl}}\downarrow$ & & $\mathbf{E}_{\textrm{key}}\downarrow$ \\ \midrule
				V2V-PoseNet~\cite{Moon2018} & 123.9 & 5.12 & 4.61 & & 137.7 & 4.85 & \textbf{3.67} & & 0.185\\
				PoseReg w/o P2PSF & 109.0 & 5.96 & 9.92 & &  114.8 & 5.53 & 8.93 & & 0.178\\
				PoseReg w/ P2PSF & 100.8 & 5.99 & 9.98 & & 108.7 & 5.54 & 8.74 & & 0.175 \\
				Ours w/o P2PSF & 98.0 & 4.96& 4.61 & & 110.8 & 4.83 & 4.39 & & 0.176\\
				Ours & \textbf{96.1} & \textbf{4.92} & \textbf{4.33} & & \textbf{108.6} & \textbf{4.77} & \underline{4.16} & & \textbf{0.173}\\
				\bottomrule
				\label{tab:quantitative}
			\end{tabular}
		\end{minipage}\hfill
		\begin{minipage}[t]{.3\linewidth}
			\centering
			\caption{Ablation study results.}
			\begin{tabular}[t]{lc}
				\toprule
				& \\
				\vspace{1.55mm}
				Method & $\mathbf{E}_{\textrm{key}}\downarrow$ \\ \midrule
				(a) w/o Poisson & 0.174\\
				(b) w/o temp. blur & 0.179\\
				(c) w/o temp. shift & 0.197\\
				(d) w/o temp. resampling & 0.185\\
				(e) All noise types & \textbf{0.173}\\
				\bottomrule
				\label{tab:ablation}
			\end{tabular}
		\end{minipage}\hfill
		\mbox{}
		\vspace{-4mm}
	\end{table*}

	\section{Experiments and Results}
	\label{sec:results}
	
	We evaluate our method against the baselines in three different experiment settings and one ablation study: (1) single subject with pseudo-transient data, using the same subject in both training and testing; (2) cross-subject with pseudo-transient data, using different subjects for training and testing; (3) cross-subject with real captured data, where we use real transient images even though our model is trained solely on synthetic data. We further conduct an ablation study to show the importance of each augmentation in our proposed pseudo-transient image synthesis procedure.
	
	\vspace{2mm}
	\noindent
	\textbf{Single-Subject Evaluation.}
	We train our model for each subject. We use a 80-20 train-test data split. 
	As shown in the quantitative results (Table~\ref{tab:quantitative} (left)), we can see our approach outperforms other baselines in terms of both pose-based metric (MPJPE) and physics-based metrics ($\mathbf{E}_{\textrm{vel}}$, $\mathbf{A}_{\textrm{accel}}$). We also present qualitative results in Fig.~\ref{fig:single_and_cross} (a). As the white dotted rectangles show, V2V-PoseNet and our method without P2PSF Net are not able to reproduce finer motion like walking. Also, non-physics-based methods (\ie, all baselines except Ours w/o P2PSF) often cause the humanoid's foot to sink into the ground (highlighted by the yellow circles). In contrast, our method produces 3D human pose closer to the ground-truth than any other baseline. 
	
	\vspace{2mm}
	\noindent
	\textbf{Cross-Subject Evaluation.}
	To further evaluate the robustness and generalization of our method, we perform cross-subject experiments where we train our models on four subjects and test on the other subject. This is challenging since different people have unique action characteristics. As shown in Table~\ref{tab:quantitative}(middle), our method once again outperforms other baselines in terms of MPJPE and $\mathbf{E}_{\textrm{vel}}$. For the smoothness metric $\mathbf{A}_{\textrm{accel}}$, V2V-PoseNet is quantitatively the best but this is because it mainly outputs a single pose only (indicated by the high MPJPE) and cannot estimate finer movement such as walking or raising hands, as shown in Fig.~\ref{fig:single_and_cross} (b). As the qualitative results in Fig.~\ref{fig:single_and_cross} (b) shows, our method with P2PSF Net produces poses closest to the ground truth, although it still fails during difficult sequences. Green arrows in Fig.~\ref{fig:single_and_cross} (b) highlight large differences in the estimated pose compared to the ground truth.
	
	\begin{figure}
		\begin{center}
			\includegraphics[width=\linewidth]{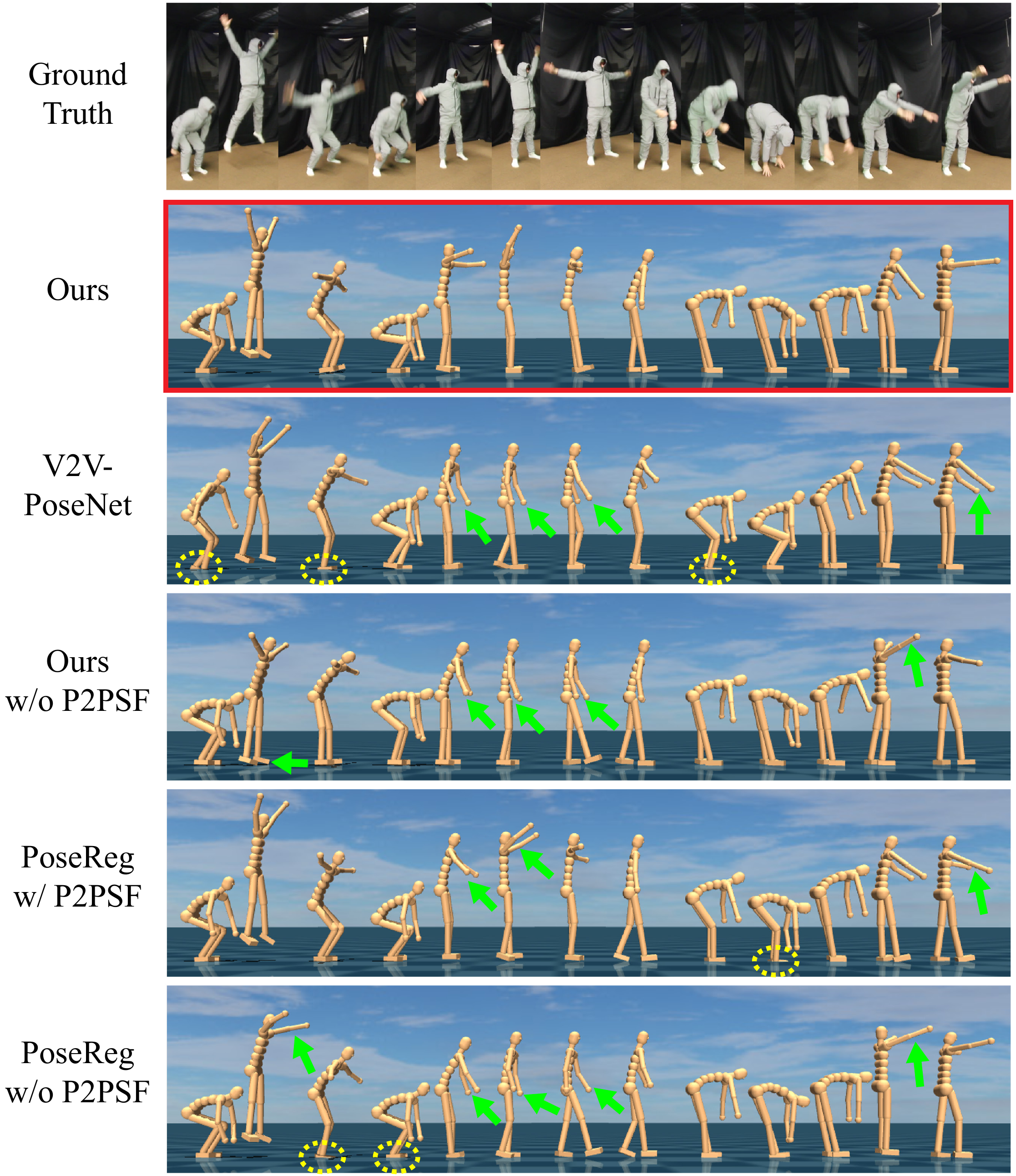}
		\end{center}
		\vspace{-2mm}
		\caption{Qualitative results for real captured transient.}
		\label{fig:stanford}
		\vspace{-3mm}
	\end{figure}
	
	\vspace{2mm}
	\noindent
	\textbf{Real Captured Cross-Subject Evaluation.}
	To show our approach's ability to work with real-world data, we further test our method on the real captured transient image dataset described in Sec.~\ref{subsec:datasets}, with our model trained with the same dataset with single subject experiment. As shown in Table~\ref{tab:quantitative} (Right), our approach produces more accurate poses than the baselines. We also present qualitative results in Fig.~\ref{fig:stanford}. As the figure shows, our method with P2PSF Net is better than the baselines in estimating human poses, including jumping (second from the left) and hand raising (sixth from the left, first from the right). Again, non-physics-based methods cause the humanoid's foot to sink into the ground (highlighted by the yellow circle).  The green arrows once again identify large discrepancies with the ground truth.
	
	\vspace{2mm}
	\noindent
	\textbf{Ablative Analysis.}
	As described in Sec.~\ref{sec:pseudo_transient}, our pseudo-transient images model the characteristics of real captured transient measurements. This ablation test investigates the effect of the four operations used to minimize the domain gap between the pseudo and real transient images: (a) Poisson noise, (b) temporal blur, (c) temporal shift, and (d) temporal re-sampling. To investigate the importance of each of them, we train our model four times under the real captured cross-subject setting, and each time exclude one of the four operations. As indicated in Table~\ref{tab:ablation}, the temporal shifts are of critical importance for performance. Temporal re-sampling also improves accuracy.  Removing Poisson noise and temporal blur independently does not severely affect performance. The combination of all four operations result in the best pose accuracy.

	\section{Conclusion}
	\label{sec:conclusion}
	
	This work brings together a diverse set of sub-areas of computer vision, including state-of-the-art in computational imaging, physics-based vision, human pose estimation and physics-based deep reinforcement learning. As a result of this integration, we have shown for the first time that it is possible to take noisy real-world photon-level measurements of human motion and transform that information into a high-level understanding of human body dynamics aided by the power of data-driven machine learning.
	
	While the primary technical focus of this work is to better understand how visual information should be represented and processed to enable 3D pose estimation from NLOS imaging, the technology described in this work also has some practical applications for next generation autonomous systems. In the context of autonomous driving, the ability to detect and track people outside of the line of sight of its sensors can be instrumental in informing planning algorithms and preventing accidents. In the context of domestic robots, the ability to see around walls could help robots make more informed decisions when entering a room or avoiding collisions. Though more research is necessary to lower the financial cost and computational complexity of the NLOS imaging system described in this work, we believe that this preliminary work shows the remarkable potential for higher-level reasoning using NLOS imaging in the real-world.

	\noindent
	\textbf{Acknowledgements.}
	We thank Ioannis Gkioulekas for many helpful suggestions. M. Isogawa is supported by NTT Corporation. M. O'Toole is supported by the DARPA REVEAL program.

	{\small
		\bibliographystyle{ieee_fullname}
		\bibliography{reference.bib}
	}

	\onecolumn
	
	\newpage
	\newpage
	
	\appendix
	\noindent
	\large
	\textbf{Appendix}
	\normalsize

	\section{Non-Line-of-Sight~(NLOS) Imaging}
	\label{sec:nlos}
	
	\subsection{The Property of Transient Images: How to Acquire, How It Looks Like, and Why the Task is Difficult}
\begin{figure}[h]
		\begin{center}
			\includegraphics[width=1.0\textwidth]{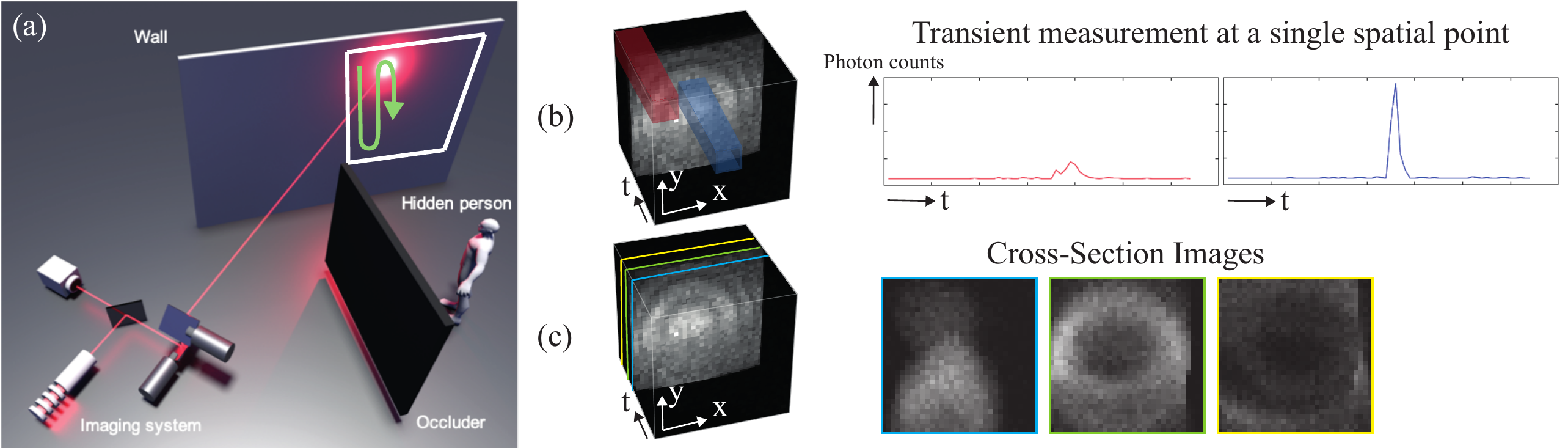}
		\end{center}
		\caption{Overview of confocal transient imaging. \textbf{(a)} A pulsed laser illuminates the wall at a point and a transient sensor measures the temporal response at the same point.  The system then raster scans the wall one point at a time to form a 3D volume of measurements. \textbf{(b}) \& \textbf{(c)} Deciphering meaningful 3D structure from transient images is non-trivial, which makes human pose estimation from transient imaging a difficult problem.}
		\label{fig:transient}
	\end{figure}

	This section aims to give a better intuition for the property of the transient images and their data acquisition processes. 
	A transient image is a 3D measurement volume, containing a scene's spatio-temporal response to laser pulse.  Each voxel encodes the number of photons at a specific (2D) point in space and at a specific (1D) point in time (see Fig.~\ref{fig:transient} (b)). In confocal NLOS imaging~\cite{o2018confocal}, the transient image captures light travelling between a specific point $(x,y)$ on a wall, and a hidden scene. As shown in Fig.~\ref{fig:transient} (a), a single pulsed laser and a transient sensor record the time light takes to travel from a point on a wall to the person hidden from the sensor's line of sight. That is, the laser light first travels (i) from the pulsed laser to the visible wall, and the visible wall to the hidden person. Then, (ii) reflected laser from the person goes back to the visible wall, and finally acquired by a co-axial transient sensor. This time of flight data acquisition is repeated for a $n \times n$ grid of point on the wall (\eg, 32$\times$32 points) by raster scanning the surface one point at a time (see green line in Fig.~\ref{fig:transient} (a)). 
	
	As shown in Fig.~\ref{fig:transient} (b), for any one point $(x,y)$, a transient measurement is a histogram of the travel time of photons. The location of the peak intensity represents the travel time required for most photons to return to point $(x,y)$, and corresponds to the distance that separates the hidden object from point $(x,y)$. The confocal transient image is a collection of such measurements for all points on the wall, and has the dimension (x,y,t) = (n,n,b), where $b$ is the number of travel time bins in the histogram.  The confocal transient images used in this work~\cite{Lindell2019} were sampled at a resolution of $32 \times 32$ spatial points and $4096$ time bins.
	Cross-sections of a single transient image at different instances in time are shown in Fig.~\ref{fig:transient} (c). The transient images show the light reflected by a hidden object, and appearing on a visible surface (\ie, wall) as a function of time.
	
\begin{figure}[h]
		\begin{center}
			\includegraphics[width=1.0\textwidth]{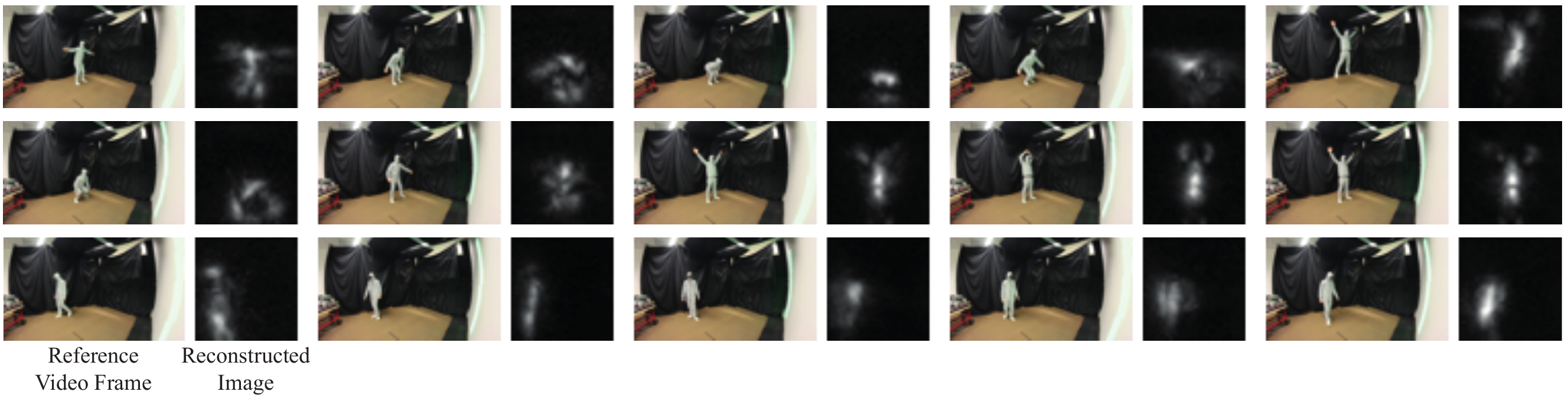}
		\end{center}
		\vspace{-3mm}
		\caption{Reconstructed human shape by NLOS imaging~\cite{o2018confocal}. Even with the state-of-the-art NLOS imaging algorithm, the reconstructed results are very blurry, noisy and have both spatially and temporally low resolution, which makes our task (\ie, physics based 3D human pose estimation) quite challenging.}
		\label{fig:lct_res}
	\end{figure}

	As made evident in Fig.~\ref{fig:transient} and \ref{fig:lct_res}, estimating physically consistent 3D human pose from transient images is challenging, when compared to working with regular RGB and RGBD data. Fig.~\ref{fig:lct_res} shows reconstructed human shape with the NLOS imaging algorithm~\cite{o2018confocal} used in our pipeline. The 2D images were generated by taking max intensity of the reconstructed 3D volume given transient images via this method.
	Even with this state-of-the-art NLOS imaging method, the reconstructed images are noisy, have low spatial resolution, and are recorded at slow frame rates. This makes it difficult to capture small shape details and fast motion, both of which are important factors when estimating a human pose sequence. Furthermore, due to the light lost after multiple scattering events, very few photons reach the sensor and the acquired transient image can therefore be very noisy. All of these characteristics make it very challenging to estimate 3D human pose directly from the transient image. Please note that due to the unique visual properties of transient images, current state-of-the-art human pose estimation methods for RGB image frames (\eg, \cite{Cao2018,Fang2017,Xiu2018}) cannot be applied directly to these transient images.
	
	\subsection{Pseudo-Transient Images}
	
	Sec.~\ref{sec:our_data_synthesis} in the main paper introduced our pseudo-transient image synthesis as a training data. We introduced four different types of noises and operations to close the domain gap that exists between pseudo-transient images and real transient images.
	This section shows visualizations of pseudo-transient images, and also provides additional details on the \emph{temporal resampling} strategy used for data augmentation.
	
	\subsubsection{Visualized Results of the Pseudo-Transient Images, and Reconstructed Depth}
\begin{figure}[h]
		\begin{center}
			\includegraphics[width=1.0\textwidth]{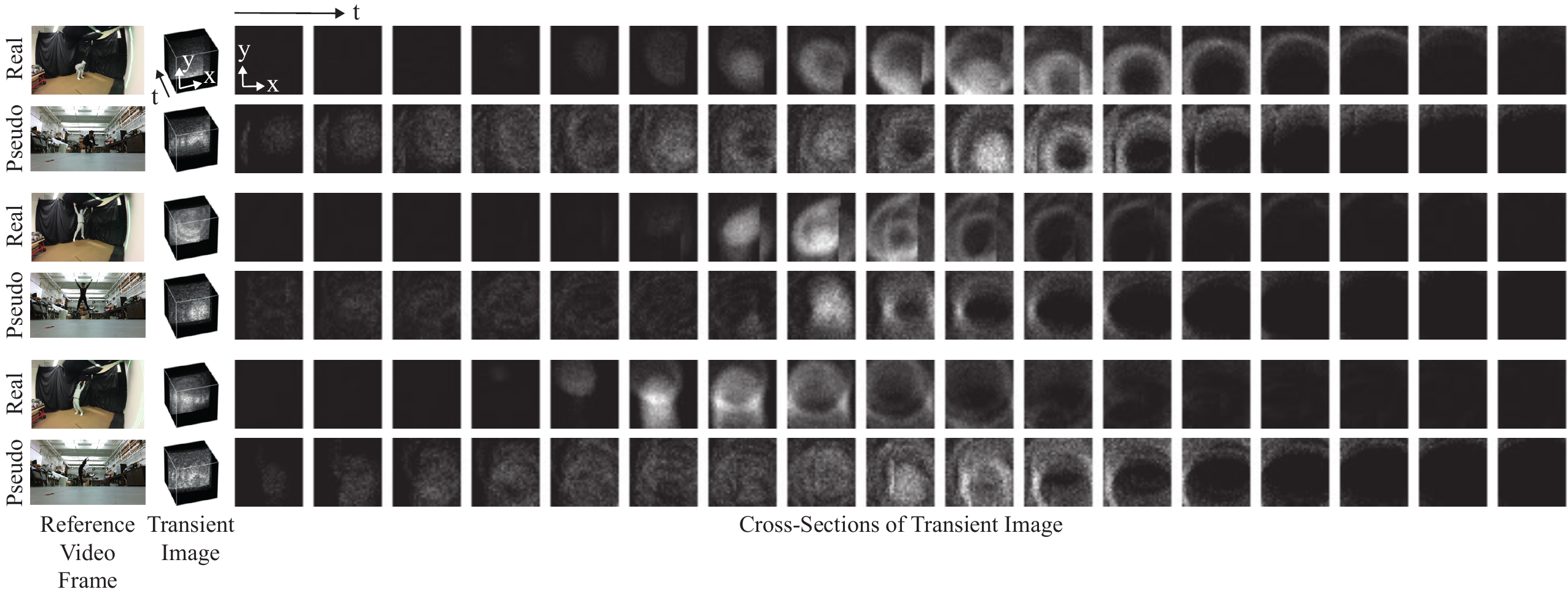}
		\end{center}
		\vspace{-3mm}
		\caption{Visualization of the real captured transient images and pseudo-transient images, for persons sharing a similar pose. 
			Note that the reference video frames are only used for comparison purposes; they are not used as input during pose estimation.}
		\label{fig:transient_comparison}
	\end{figure}

	Fig.~\ref{fig:transient_comparison} highlights both synthesized pseudo-transient images and real transient images of persons sharing a similar pose. To better visualize the volumes, we also show cross-sections (in time) of these transient images. Note that the cross-section containing the highest signal occurs at different times, because this depends on the person's location in the hidden environment. We apply five levels of temporal shift (see the main paper Sec.~\ref{sec:our_data_synthesis}) to augment the pseudo-transient images, and make our procedure robust to a person's position. Some of the slices also show discontinuities along the horizontal axis; this is an artifact of raster scanning a wall while a person moves within the hidden scene.  We reproduce these artifacts for our pseudo-transient images by simulating the same raster scanning procedure; please refer Sec.~\ref{sec:augmentation_strategy} for more details. Note that the reference video frames are only for reference, and not used in the pose estimation process.  As shown in the figure, our pseudo-transient images closely match the real transient images.

	\subsubsection{Data Augmentation Strategy (Temporal Resampling)}
	\label{sec:augmentation_strategy}
	
	\begin{figure}[ht!]
		\begin{center}
			\includegraphics[width=1.0\textwidth]{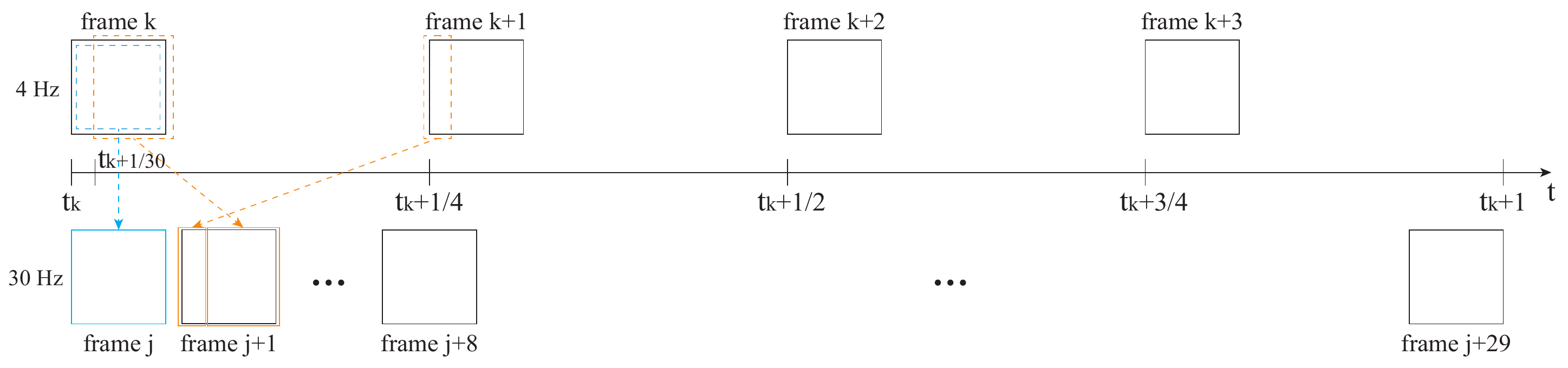}
		\end{center}
		\caption{Temporal resampling from 4 Hz to 30 Hz.}
		\label{fig:temporal_resampling}
	\end{figure}
	
	Confocal NLOS imaging requires raster scanning a visible surface point by point. For the data used in this work~\cite{Lindell2019}, the confocal transient measurements are sampled at $32 \times 32$ locations at a frame rate of $4$ Hz.

	As discussed in Sec.~\ref{sec:our_data_synthesis}, we introduce a procedure to temporally re-sample transient image sequences.  First, to simulate the raster scanning procedure, we capture depth maps at $30$ Hz, convert them to pseudo-transient measurements, and simulate raster-scanned measurements by down-sampling the result to $4$ Hz.  Second, we invert the raster-scanning process, and up-sample the pseudo-transient data from $4$ Hz back to $30$ Hz.
	
	As shown in the top row of Fig.~\ref{fig:temporal_resampling}, each transient image is a collection of transients scanned from time $t_k$ to $t_k + \frac{1}{4}$, where $t_k = \frac{k}{4}$ represents the start time of the $k^\text{th}$ frame (in seconds). 
	To generate a $30$ Hz transient sequence (the bottom row), we simply assemble transients captured within the same time range, but set the start time of the $k^\text{th}$ frame to $t_k = \frac{k}{30}$.
	
	Here, to simplify the explanation, in Fig.~\ref{fig:temporal_resampling} we assume $j^\text{th}$ frame in $30$ Hz transient sequence starts at the same start time with $k^\text{th}$ frame in $4$ Hz. 
	Assembling $30$ Hz frames is simply combining $4$ Hz frames with starting scanning point $t_k + \beta$, where $\beta>0$ incremented with a constant step $\frac{1}{30}$ sec. For example, since the starting time of $k^\text{th}$ frame in $4$ Hz $j^\text{th}$ frame in $30$ Hz is the same, $j^\text{th}$ frame in $30$ Hz is obtained by just copying $k^\text{th}$ frame in $4$ Hz as the blue lines show. The next $j+1^\text{th}$ frame in $30$ Hz is assembled with $4$ Hz frame scanned from $t_k + \frac{1}{30}$ to $t_k + \frac{1}{4} + \frac{1}{30}$. Thus as orange lines show, copying scanned data in this time range to the same spatial point generates resampled frame $j+1^\text{th}$.
	Converting frames from $30$ Hz to $4$ Hz is just a inversion of this process.

	\subsection{Background: Detailed Derivation of the Confocal NLOS Imaging}
	\label{sec:confocal_nlos}
	
	The confocal NLOS imaging that we use~\cite{o2018confocal} aims to convert 3D transient image $\tau(x',y',t)$ into a discretized reconstruction 3D volume $\rho(x, y, z)$, which represents the reflectivity at every point $(x,y,z)$, $z>0$ in space, as we show with Eq.~\textcolor{blue}{3} in the main paper. Here, this section shows more detailed derivation of the equation.
	
	In the continuous domain, the relationship between a 3D reconstruction volume $\rho(x, y, z)$ and a 3D transient image $\tau(x, y, t)$ are represented as below.
	\begin{equation}
	\begin{array}{l}
	\tau(x',y',t) \quad=\quad
	\iiint_{\Omega} \tfrac{1}{r^4}~\rho(x,y,z)~\delta(2 \sqrt{(x'-x)^2+(y'-y)^2+z^2}-tc)~dx~dy~dz,
	\end{array}
	\label{eq:appendix_lct_1}
	\end{equation}
	where $c$ is the speed of light. Eq.~\ref{eq:appendix_lct_1} shows that a transient measurement sample $\tau(x',y',t)$ captures the photon flux at point $(x',y')$ and time $t$, relative to an pulse scattered by the same point at time $t=0$. $\Omega$ represents a 3D half-space containing the hidden space on one side of the wall ($z>0$). The Dirac delta function $\delta$ represents a four-dimensional spatio-temporal hypercone surface, represented by $x^2+y^2+z^2-(tc/2)^2$; this hypercone (or light-cone) models light propagation from the visible wall to the object and back to the visible wall.
	Here, the distance function $r = \sqrt{(x'-x)^2+(y'-y)^2+z^2} = tc/2$ in Eq.~\ref{eq:appendix_lct_1} can be represented with the light arrival time $t$. Thus, the term $1/r^4$ can be removed from the triple integral. Also, by replacing variables as $z = \sqrt{u}$, $dz/du = 1/(2 \sqrt{u})$, and $v = (tc/2)^2$, the Equation \ref{eq:appendix_lct_1} can be re-written as
	\begin{equation}
	\begin{array}{l}
	\underbrace{v^{3/2}\tau(x',y',2\sqrt{v}/c)}_{R_t\{\tau\}(x',y',v)} \quad=\quad
	\iiint_{\Omega} \underbrace{\tfrac{1}{2\sqrt{u}} ~\rho(x,y,\sqrt{u})}_{R_z\{\rho\}(x,y,u)} ~\underbrace{\delta((x'-x)^2+(y'-y)^2+z^2+u-v)}_{h(x'-x, y'-y, v-u)} ~dx~dy~du \,,
	\end{array}
	\label{eq:appendix_lct_2}
	\end{equation}
	which can be expressed as a 3D convolution $R_t\{\tau\} = h \ast R_z\{\rho\}$, where $\ast$ is the 3D convolution operator. $h$ is a known 3D point spread function (PSF), and describes the transient response of a single scatterer. $R_z\{\cdot\}$ resamples $\rho$ along the z-axis and attenuates the result by $1/2\sqrt{u}$, and $R_t\{\cdot\}$ resamples $\tau$ along the time axis and scales the result by $v^{\frac{3}{2}}$.
	With resampled transient image and reconstruction volume as $\Tilde{\tau}=R_t\{\tau\}$ and $\Tilde{\rho}=R_z\{\rho\}$ respectively, the forward image formation model for confocal NLOS images becomes a simple 3D convolution operation:
	\begin{equation}
	\Tilde{\tau} = h * \Tilde{\rho}\,,
	\end{equation}
	Also, this convolution can be rewritten with the matrix-form as follows:
	\begin{equation}
	\tau = 
	R_t^{-1} F^{-1} \hat{H} F R_z \rho\,,
	\label{eq:lct_discrete_appendix}
	\end{equation}
	where we vectorize volumes $\tau$ and $\rho$.  The matrix $F$ represents a 3D discrete Fourier transform and $\hat{H}$ is a diagonal matrix representing the Fourier transform of the PSF $h$.
	The NLOS imaging process reconstructs a 3D volume $\rho_\ast$ from a transient image $\tau$ by inverting Eq.~\ref{eq:lct_discrete_appendix} and solving a 3D deconvolution procedure (\eg, using the Wiener filter):
	\begin{equation}
	\rho_{*} = 
	R_z^{-1} F^{-1} \underbrace{\left[\frac{\hat{H}^*}{|\hat{H}|^2 + \frac{1}{\alpha}}\right]}_{\textbf{the inverse PSF}} F R_t \tau\,,
	\label{eq:lct_final_appendix}
	\end{equation}
	where a user-defined parameter $\alpha$ controls how sensitive the inverse PSF is to noise.

	\section{DeepRL Based Pose Estimation Pipeline}
	\label{sec:deep_rl}
	
	This section explains implementation details about our DeepRL based 3D human pose estimation pipeline (Sec.~\textcolor{blue}{3.1} in the main paper). 
	We use a humanoid model and a physics simulator, and formalize our task of estimating a pose sequence $p_{1:T}$ from a transient image sequence $\tau_{1:T}$ with a Markov Decision process~(MDP). The MDP is defined by a tuple $\mathcal{M} = (S, A, P, R, \gamma)$ of states, actions, transition dynamics, a reward function, and a discount factor.
	At each time step, the humanoid agent samples an action $a_t$ from a policy $\pi (a_t | s_t)$ whose input state $s_t$ contains both the visual context $\phi_t$ (computed from the transient images) and the humanoid state $z_t$ (\emph{i.e.}, joint angles and velocities). Next, the environment generates the next state $s_{t+1}$ through physics simulation and gives the agent a reward $r_t$ based on how well the humanoid's 3D pose aligns with the ground-truth. 
	Detailed definitions of the state $s_t$, action $a_t$, and policy $\pi_\theta$ are given in the main paper.
	
	To solve this MDP, inspired by previous works~\cite{Peng2018_1,Yuan2019}, we apply the Proximal Policy Optimization~(PPO)~\cite{Schulman2017} algorithm to obtain the optimal policy $\pi^*$ that maximizes the expected discounted return $\mathbb{E} [ \sum^T_{t=1} \gamma^{t-1} r_t ]$. In the following, we first describe the design of the reward function $r_t$ in Sec.~\ref{sec:reward}, and then explain the policy learning algorithm in Sec.~\ref{sec:policy_learning}. Additionally, we describe in Sec.~\ref{sec:failsafe} the fail-safe mechanism we use to help the humanoid recover from unstable states.
	
	\subsection{Reward Function}
	\label{sec:reward}
	
	This section describes the specific reward function we use to train the humanoid policy. 
	Following~\cite{Yuan2019}, to encourage the policy to output a pose sequence $p_{1:T}$ that matches the ground-truth $\hat{p}_{1:T}$,  we define the reward function as
	
	\begin{equation}
	r_t = w_q r_q + w_e r_e + w_p r_p + w_v r_v,
	\end{equation}
	where $w_p, w_v, w_q, w_e$ are weighting factors. 
	
	The pose reward $r_q$ measures the difference between pose $p_t$ and the ground-truth $\hat{p}_t$ for non-root joints. Let $q^j_t$ and $\hat{q}^j_t$ denote the $j$-th joint's orientation quaternion of the estimated pose and the ground-truth pose respectively. The pose reward $r_q$ is computed as 
	\begin{equation}
	r_{q}=\exp \left[-2\sum_{j}\left\|q_{t}^{j} \ominus \hat{q}_{t}^{j}\right\|^{2}\right].
	\end{equation}
	
	The end-effector reward $r_e$ evaluates the difference between the local vector of end effector $e_t$ and the ground-truth $\hat{e}_t$ . We use head, hands, and feet as end-effectors. The end-effector reward $r_e$ is defined as
	\begin{equation}
	r_{e}=\exp \left[-20\sum_{e}\left\|e_{t}-\hat{e}_{t}\right\|^{2}\right].
	\end{equation}
	
	The root pose reward $r_p$ encourages the humanoid’s root joint to have the same height $h_t$ and orientation quaternion $q^r_t$ as the ground-truth $\hat{h}_t$ and $\hat{q}^r_t$:
	\begin{equation}
	r_{p}=\exp \left[-300\left(\left(h_{t}-\hat{h}_{t}\right)^{2}+\left\|q_{t}^{r} \ominus \hat{q}_{t}^{r}\right\|^{2}\right)\right].
	\end{equation}
	
	The root velocity reward $r_v$ penalizes the deviation of the root's linear velocity $l_t$ and angular velocity $\omega^r_t$ from the ground-truth $\hat{l}_t$ and $\hat{w}_t$:
	\begin{equation}
	r_{v}=\exp \left[-\left\|l_{t}-\hat{l}_{t}\right\|^{2}-0.1\left\|\omega_{t}^{r}-\hat{\omega}_{t}^{r}\right\|^{2}\right].
	\end{equation}
	
	\subsection{Policy Learning}
	\label{sec:policy_learning}
	To compute the optimal humanoid policy $\pi^*$ that maximizes the expected discounted return, we use the Proximal Policy Optimization~(PPO)~\cite{Schulman2017} algorithm to compute the gradients of the policy $\pi_\theta$. Traditional policy gradient methods often suffer from catastrophic failure (\ie, the policy $\pi_\theta$ becomes irrecoverably bad) due to noisy policy gradients caused by high variance of the data collected by the policy. To address this problem, PPO uses a mechanism that prevents noisy gradients from changing the policy too much. Specifically, PPO utilizes a clipping function $clip\left(w_{t}(\theta), 1-\epsilon, 1+\epsilon\right)$ that clips a gradient by setting it to zero whenever the ratio of current/old policies $w_{t}(\theta)$ is more than $\epsilon$ away from $1$. Then, PPO minimizes the following loss function:
	\begin{equation}
	L(\theta)=\mathrm{E}_{s_{t}, a_{t}}\left[\min \left(w_{t}(\theta) \mathcal{A}_{t}, \operatorname{clip}\left(w_{t}(\theta), 1-\epsilon, 1+\epsilon\right) \mathcal{A}_{t}\right)\right],
	\label{eq:ppo1}
	\end{equation}
	\begin{equation}
	w_{t}(\theta)=\frac{\pi_{\theta}\left(a_{t} | s_{t}\right)}{\pi_{\theta_{old}}\left(a_{t} | s_{t}\right)}\,,
	\label{eq:ppo2}
	\end{equation}
	where $\mathcal{A}_t$ is an advantage estimate that measures the goodness of taking action $a_t$ in state $s_t$ and $\pi_{\theta_{old}}$ is the old policy before the gradient update with fixed parameters $\theta_{old}$.
	Our PPO-based policy learning procedure is outlined in Algorithm~\ref{alg:policy}.

	\begin{algorithm}
		\caption{Policy Learning}
		\label{alg:policy}
		\begin{algorithmic}[1]
			\State \textbf{Output:} parameters $\theta$ of the optimal policy $\pi^\ast$
			\State Initialize $\theta$ randomly
			\While{not converged}
			\State $s_0$ $\leftarrow$ sample initial state from the ground truth motion
			\For{ each simulation step $t$}
			\State $s_t$ $\leftarrow (\phi_t, z_t)$ 
			\State $a_t \sim \pi_{\theta}(a_t | s_t)$
			\State Apply $a$ and simulate forward one step
			\State $s_{t+1}$ $\leftarrow$ new state from simulation
			\State $r_t$ $\leftarrow$ pose matching reward  \Comment{Reward function: Sec.~\ref{sec:reward}}
			\State Record $(s_t, a_t, r_t, s_{t+1})$ into a memory
			\EndFor
			\State $\theta_{old}$ $\leftarrow$ $\theta$
			\For{ each update step }
			\State Sample mini-batch of $n$ samples {$(s_i, a_i, r_i, s_{i+1})$} from a memory
			\For{ each $(s_i, a_i, r_i, s_{i+1})$ }
			\State $\mathcal{A}_i$ $\leftarrow$ compute advantage~\cite{Schulman2017}
			\State $w_{i}(\theta)$ $\leftarrow$ $\frac{\pi_{\theta}\left(a_{i} | s_{i}\right)}{\pi_{\theta_{old}}\left(a_{i} | s_{i}\right)}$ \Comment{Eq.~\ref{eq:ppo2}}
			\EndFor
			\State $\theta$ $\leftarrow$ 
			$\theta + \frac{1}{n} \Sigma_{i} 
			\nabla_{\theta} \min (w_{i}(\theta) \mathcal{A}_{i}, \operatorname{clip}\left(w_{i}(\theta), 1-\epsilon, 1+\epsilon\right) \mathcal{A}_{i})$ \Comment{Eq.~\ref{eq:ppo1}}
			\EndFor
			\EndWhile
		\end{algorithmic}
	\end{algorithm}

	\subsection{Fail-safe Mechanism and the state regressor $\mathcal{F}$}
	\label{sec:failsafe}
	Sometimes the extreme noise in the transient images can produce irregular control which makes the humanoid fall down to the ground. To prevent this problem, we use the same value function-based fail-safe mechanism as described in \cite{Yuan2018} to detect unstable humanoid states and reset the humanoid state to the output of the state regressor $\mathcal{F}$. Without using any physics simulation, the state regressor $\mathcal{F}$ directly maps the visual context $\phi_t$ to the corresponding state $z_t$ with an MLP of two hidden layers (300, 200) and ReLU activations. Using supervised learning, we train $\mathcal{F}$ for 100 iterations with Adam~\cite{adam} and a learning rate of 1e-4.
	
	\section{P2PSF Net Architecture}
	
\begin{figure}[h]
		\vspace{-2mm}
		\begin{center}
			\includegraphics[width=1.0\textwidth]{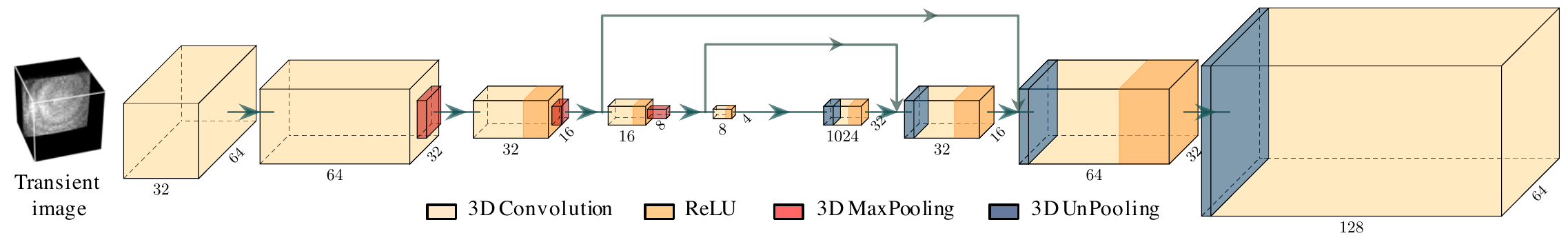}
		\end{center}
		\vspace{-3mm}
		\caption{The architecture of P2PSF Net.}
		\label{fig:p2psf}
		\vspace{-8mm}
	\end{figure}

	We show in Fig.~\ref{fig:p2psf} a more detailed version of the P2PSF Net (Fig.~\ref{fig:models} in the main paper).
	As discussed in the main paper (Sec.~\ref{sec:transient_to_pose}), P2PSF Net is a volume to volume network introduced as our feature extractor (the network that obtains the transient feature $\psi_t$ from an transient image $\tau_t$). P2PSF Net takes a transient image of resolution $(x,y,t)=32\times32\times64$ and outputs a volume $(x,y,d)=128\times64\times64$, matching the size of the inverse PSF volume used in confocal NLOS imaging (Eq.~\ref{eq:lct_final_appendix}). The P2PSF Net has nine 3D convolution layers and two residual connections. Please refer to Fig.~\ref{fig:p2psf} for the specific size of each layer.
	
\end{document}